%% file: root.tex
\documentclass[letterpaper, 10 pt, conference]{ieeeconf}  

\IEEEoverridecommandlockouts                              

\usepackage{cite}
\usepackage{xcolor}
\usepackage{amsmath,amssymb,amsfonts}
\usepackage{graphicx}
\usepackage{textcomp}
\usepackage{bm}
\usepackage{algorithm}
\usepackage{algpseudocode}
\usepackage{tabularx}
\usepackage{multirow}
\usepackage{booktabs}
\newcommand{\tablefontsize}{\scriptsize}
\newcolumntype{P}[1]{>{\centering\arraybackslash}p{#1}}

\DeclareMathOperator*{\minimize}{min}

\newcommand{\svector}[1]{\boldsymbol{\mathrm{#1}}}
\newcommand{\smatrix}[1]{\boldsymbol{\mathrm{#1}}}
\newcommand{\red}[1]{{\color{red}#1}}
\newcommand{\blue}[1]{{\color{blue}#1}}

\title{\LARGE \bf
Robustified Time-optimal Point-to-point Motion Planning and Control under Uncertainty*
}

\author{Shuhao Zhang$^{1}$ and Jan Swevers$^{1}$
\thanks{*This work has been carried out within the European Union’s Horizon 2020 research and innovation programme under the Marie Skłodowska-Curie grant agreement No.953348 ELO-X.}
\thanks{$^{1}$Shuhao Zhang and Jan Swevers are with the MECO Research Team, the Department of Mechanical Engineering, KU Leuven, and Flanders Make@KU Leuven, Core lab MPRO, 3001, Leuven, Belgium (e-mail: {\tt\small firstname.lastname@kuleuven.be}).}%
}

\begin{document}

\maketitle
\thispagestyle{empty}
\pagestyle{empty}

\begin{abstract}
    
This paper proposes a novel approach to formulate time-optimal point-to-point motion planning and control under uncertainty. 
The approach defines a robustified two-stage Optimal Control Problem (OCP), in which stage 1, with a fixed time grid, is seamlessly stitched with stage 2, which features a variable time grid.
Stage 1 optimizes not only the nominal trajectory, but also feedback gains and corresponding state covariances, which robustify constraints in both stages. 
The outcome is a minimized uncertainty in stage 1 and a minimized total motion time for stage 2, both contributing to the time optimality and safety of the total motion. 
A timely replanning strategy is employed to handle changes in constraints and maintain feasibility, while a tailored iterative algorithm is proposed for efficient, real-time OCP execution.

\end{abstract}
\input{SecI}

\input{SecII}
\input{SecIII}

\input{SecIV}
\input{SecV}


\bibliographystyle{IEEEtran}
\bibliography{reference}

\end{document}

%% file: SecI.tex
\section{Introduction}
Time-optimal motion planning and control has garnered significant attention in applications such as racing, search and rescue, and other time-critical scenarios.
In racing applications, Model Predictive Contouring Control (MPCC), which maximizes the tracking progress along a known trajectory to achieve the time-optimal objective, is considered a promising approach \cite{Denise_10, Juraj_19, Krinner_24}.
MPCC emphasizes real-time trajectory tracking of the system rather than motion and trajectory planning.
In contrast, applications including search and rescue \cite{Hazim_19}, robotic manipulators \cite{Geering_86} and cranes \cite{Verschueren_17} require addressing time-optimal point-to-point motion planning and control, which is the focus of this paper.

Time-optimal point-to-point motion planning can be directly approached in the system’s state space as a discrete-time Optimal Control Problem (OCP) formulated using methods such as exponential weighting \cite{Verschueren_17}, time scaling \cite{Rösmann_15}, or a two-stage method \cite{ZSH_24_2s}, which combines aspects of both.
Due to uncertainties such as process noise, the system may face challenges in accurately tracking the planned motion trajectory if these uncertainties are not considered in the OCP. 
In such cases, an additional low-level feedback tracking controller becomes essential. 
Additionally, the OCP can be repeatedly solved with updated information on current state using a Nonlinear Model Predictive Control (NMPC) scheme \cite{MPC}.
Given that the optimal solution of the time-optimal OCP often lies on the edge of stage constraints, a straightforward approach is to introduce fixed, heuristically chosen safety margins to account for potential constraint violations due to uncertainties in actual motions. 
However, this approach lacks formal safety guarantees and may result in overly conservative motions.

Safety margins derived from ellipsoidal state uncertainty sets, approximated through linearization-based covariance propagation, are commonly employed to robustify stage constraints by defining chance constraints in robust \cite{Houska_10} and stochastic \cite{Gillis_13, Hewing_20} NMPC problems.
This inevitably leads to a significant increase in the number of decision variables, consequently escalating the computational complexity.
Nevertheless, tailored iterative algorithms \cite{Feng_20, Zanelli_21, Messerer_21, Gao_23} can efficiently obtain sub-optimal solutions by decoupling the uncertain components from the robustified problem.
These tailored algorithms are suitable for solving the robustified time-optimal OCP formulated using the exponential weighting method because a fixed time grid is employed, although their efficacy may diminish when the number of control steps is large.
The robustified time-optimal OCP using time scaling approach offers a more straightforward formulation to achieve general long-horizon planning, with \cite{ZSH_24} proposing a tailored algorithm combined with precomputed feedback gains to solve it.
However, due to the variable sparse time grid defined by time scaling, the resulting trajectory must be interpolated to align with the system’s control frequency. This interpolation makes it impossible to ensure that the whole refined trajectory satisfies the constraints, which can lead to infeasibility during replanning.

In this paper, we propose a robust two-stage time-optimal Optimal Control Problem (OCP) outlined in Sec.\ref{Sec:II}. 
In stage 1, a fixed time grid is employed, and the optimized variables include a nominal time-optimal state and control trajectory, along with feedback gains designed to counteract the growth of state uncertainty. 
These components can be directly implemented on the discrete-time nonlinear motion system. 
Stage 2 incorporates a variable time grid, formulated using the time scaling method, and is seamlessly stitched with stage 1. 
The feedback gains and corresponding state covariances from stage 1 are utilized to derive safety margins that robustify the constraints of both stages. 
The objective is to achieve point-to-point motion in the shortest possible time, with the objective function minimizing the total time of stage 2 and state and control uncertainties of stage 1.
We repeatedly solve this OCP by employing an asynchronous NMPC scheme \cite{ASAP-MPC} to ensure timely replanning with full convergence under fluctuating computation delays.
In Sec.\ref{Sec:III}, we propose a tailored iterative algorithm inspired by \cite{Messerer_21} to efficiently solve this robustified OCP by decoupling it into two subproblems: a Riccati recursion to derive feedback gains and a nominal time-optimal OCP with variable safety margins, which are solved alternately.
Sec.\ref{Sec:IV} presents and discusses a numerical example, and Sec.\ref{Sec:V} concludes the paper.

\noindent\textit{Notation:} For vectors $s\in\mathbb{R}^{n}$, $u\in\mathbb{R}^{m}$, we denote their vertical concatenation by $\texttt{vec}(s,u)\in\mathbb{R}^{n+m}:=[s^\top, u^\top]^\top$. 
For a matrix $\Sigma\in\mathbb{R}^{n\times n}$, we denote its vertical concatenation along columns as $\texttt{vec}(\Sigma)\in\mathbb{R}^{n^2}$.
The L1-norm of $s$ is denoted by $\|s\|_1$.
$\texttt{diag}(s)$ denotes a diagonal matrix with $s$ as its diagonal.
$I_n$ denotes a $n\times n$ identity matrix.
The operation $\texttt{tr}(\Sigma):=\sum_{i=1}^n\Sigma_{ii}$ represents the trace of $\Sigma$.

%% file: SecII.tex
\section{Robustified time-optimal motion planning and control problem}\label{Sec:II}
Considering a discrete-time nonlinear motion system under uncertainties of the form
\begin{equation}
    s_{n+1} = f_d(s_n, u_n, w_n), 
    \label{eq: discrete_time_sys}
\end{equation}
where $s_n\in\mathbb{R}^{n_s}$, $u_n\in\mathbb{R}^{n_u}$, and $w_n\in\mathbb{R}^{n_s}$
are the system's states, controls, and uncertainties, respectively.
The uncertainties $w_n\sim\mathcal{N}(0,\Sigma_w)$ are drawn from a zero-mean Gaussian distribution with covariance $\Sigma_w$.

This paper addresses the problem of point-to-point motion planning and control under uncertainties, aiming to plan time-optimal motions for the system (\ref{eq: discrete_time_sys}) from an initial nominal state $s_{\text{t0}}$ with known associated state uncertainty $\Sigma_{\text{t0}}$
to a desired terminal state $s_{\text{tf}}$.
The objective is to minimize both state and control uncertainties, as well as the total motion time, while satisfying stage and terminal constraints.
The planned motion includes a time-optimal nominal trajectory, e.g., with length $N$ denoted by $\Bar{\svector{s}} = [\Bar{s}_{0},...,\Bar{s}_{N}]$, $\Bar{\svector{u}} = [\Bar{u}_{0},...,\Bar{u}_{N-1}]$, along with corresponding feedback gains 
$\smatrix{K} = [K_{0},...,K_{N-1}]$.
In the following, we detail all necessary elements—feedback control law, uncertainty propagation, and constraint robustification—required to formulate the proposed robustified time-optimal motion planning OCP, along with our asynchronous strategy for achieving timely replanning.
\subsection{Feedback control law and uncertainty propagation}\label{Sec:IIA}
The feedback aims at regulating deviations of the  actual state from the nominal state corresponding to the planned time optimal motion through the piecewise linear control law
\begin{equation}
    u(t) = \Bar{u}_n+K_n(s_n-\Bar{s}_n), t\in[t_n, t_{n+1}] 
    \label{eq: control_policy}
\end{equation}
with sampling time $t_s$. 
This feedback action helps to mitigate the growth of state uncertainty, which is propagated based on the system (\ref{eq: discrete_time_sys}) linearized around the nominal trajectory in the form of ellipsoidal tubes.
As detailed in \cite{Houska_10} and \cite{Gillis_13},
starting from $\Sigma_{\text{t0}}$, the state covariance used to construct ellipsoidal uncertainty tubes is propagated as follows:
\begin{equation}
    \begin{aligned}
            \Sigma_{n+1}&=(A_n+B_nK_n)\Sigma_{n}(A_n+B_nK_n)^\top+G_n\Sigma_wG_n^\top\\
            &=:\Phi\left(\Bar{s}_n, \Bar{u}_n, K_n, \Sigma_{n}\right),
    \end{aligned}
\label{eq: uncertainty_propagation}
\end{equation}
with $A_n:=\frac{\partial f_d(\Bar{s}_n, \Bar{u}_n, 0)}{\partial s_n}$, $B_n:=\frac{\partial f_d(\Bar{s}_n, \Bar{u}_n, 0)}{\partial u_n}$, and $G_n:=\frac{\partial f_d(\Bar{s}_n, \Bar{u}_n, 0)}{\partial w_n}$.
\subsection{Constraints robustification}\label{Sec:IIC}
During the motion, stage constraints $h(s_n, u_n)\in\mathbb{R}^{n_h} \leq 0$ and terminal constraints $h_{\text{tf}}(s_N)\in\mathbb{R}^{n_{h_{tf}}}\leq 0$ must be satisfied in both actual states and controls to ensure safety and achieve time optimality.
Specifically, the system must avoid collisions with obstacles and closely track the planned time-optimal trajectory without experiencing control saturation.
Therefore, we robustify each stage and terminal constraint by explicitly incorporating state covariances and feedback gains following the approach of  \cite{Gillis_15_phd}:
\begin{equation}
    \begin{aligned}
        h_i(s_n, u_n)\approx h_i(\Bar{s}_n, \Bar{u}_n)+\sigma\sqrt{\beta_{n,i}}&\leq0,\ i\in[1,n_h],\\
        h_{\text{tf},i}(s_N)\approx h_{\text{tf},i}(\Bar{s}_N)+\sigma\sqrt{\beta_{\text{tf},i}}&\leq0,\ i\in[1,n_{h_{\text{tf}}}],
    \end{aligned}
    \label{eq: robustified_constr}
\end{equation}
where factor $\sigma$ satisfies $1-\mathcal{C}(\sigma)=p$ with $p$ representing the chosen probability level for satisfying the constraints given the unbounded nature of uncertainties, and $\mathcal{C}$ denoting the cumulative distribution function for a Gaussian distribution.
Corresponding constraint variances, $\beta_{n,i}$ and $\beta_{\text{tf},i}$, are approximated based on linearization of $h_i(s_n, u_n)$ around $(\bar{s}_n, \bar{u}_n)$ and linearization of $h_{\text{tf}}(s_N)$ around $\bar{s}_N$ respectively:
\begin{equation}
    \begin{aligned}
        \beta_{n,i}
        &=\frac{\partial h_i(\bar{s}_n, \bar{u}_n)}{\partial \texttt{vec}(s_n, u_n)}\begin{bmatrix}I_{n_s}\\K_n\end{bmatrix}\Sigma_n\left(\frac{\partial h_i(\bar{s}_n, \bar{u}_n)}{\partial \texttt{vec}(s_n, u_n)}\begin{bmatrix}I_{n_s}\\K_n\end{bmatrix}\right)^\top\\
        &=:H_i(\bar{s}_n, \bar{u}_n, \Sigma_n, K_n),\\
        \beta_{\text{tf},i}
        &=\frac{\partial h_{\text{tf},i}(\bar{s}_N)}{\partial s_N}\Sigma_N\frac{\partial h_{\text{tf},i}(\bar{s}_N)}{\partial s_N}^\top
        =:H_{\text{tf},i}(\bar{s}_N, \Sigma_N).
    \end{aligned}
    \label{eq: constrs_var}
\end{equation}
\subsection{Robustified two-stage time-optimal OCP}\label{Sec:IIC}
We build upon our previous work \cite{ZSH_24_2s}, which introduced a two-stage OCP for time-optimal motion planning. 
To enhance safety under uncertainty, we now integrate both motion planning and control, formulating the following robustified two-stage time-optimal OCP:
\begin{subequations} 
    \label{ocp: time_optimal_2s_robustified}
    \begin{alignat}{3}
        \minimize_{\substack{
        \Bar{\svector{s}}_1, \Bar{\svector{u}}_1, \Bar{\svector{s}}_2, \Bar{\svector{u}}_2,\\
        \smatrix{K}_1,\smatrix{\Sigma}_1, 
        \svector{\beta}_1, \svector{\beta}_2, T_2
        }} 
        &T_2 + \sum_{n=0}^{N_1-1}l_{\Sigma}(\Sigma_{1,n}, K_{1,n})+ l_{\Sigma_{\text{tf}}}(\Sigma_{1,N_1})\label{ocp1:obj}\\
        \text{s.t.}\hspace{1em}
        \Bar{s}_{1,0}   =&\  s_{\text{t0}}, \Sigma_{1,0} = \Sigma_{\text{t0}},\label{ocp1:s1_1}\\
        \Bar{s}_{1,n+1} =&\ f_d(\Bar{s}_{1,n},\Bar{u}_{1,n}),\label{ocp1:s1_2}\\
        \Sigma_{1,n+1}  =&\ \Phi(\Bar{s}_{1,n},\Bar{u}_{1,n}, K_{1,n}, \Sigma_{1,n}),\label{ocp1:s1_3}\\
        h(\Bar{s}_{1,n}, \Bar{u}_{1,n}) &+ \sigma\sqrt{\beta_{1,n}+\epsilon}\leq0, \label{ocp1:s1_4}\\
        \beta_{1,n}     =&\ H(\Bar{s}_{1,n},\Bar{u}_{1,n}, K_{1,n}, \Sigma_{1,n}),\label{ocp1:s1_5}\\
        \Bar{s}_{2,0}   =&\  \Bar{s}_{1,N_1},\label{ocp1:s1_s2}\\
        \Bar{s}_{2,n+1} =&\ f_T(\Bar{s}_{2,n},\Bar{s}_{2,n},T_2/N_2),\label{ocp1:s2_1}\\
        h(\Bar{s}_{2,n},\Bar{u}_{2,n}) &+ \sigma\sqrt{\beta_{2,n}+\epsilon}\leq0,\label{ocp1:s2_2}\\ 
        \beta_{2,n} =&\ H(\Bar{s}_{2,n},\Bar{u}_{2,n}, K_{1,N_1-1}, \Sigma_{1,N_1-1}),\label{ocp1:s2_3}\\
        h_{\text{tf}}(\Bar{s}_{2,N_2}) &+\sigma\sqrt{\beta_{2,N_2} +\epsilon}\leq0,\label{ocp1:s2_4}\\
        \beta_{2,N_2}   =&\ H_{\text{tf}}(\Bar{s}_{2,N_2}, \Sigma_{1,N_1}),\label{ocp1:s2_5}\\
        \Bar{s}_{2,N_2}   =&\  s_{\text{tf}} ,\label{ocp1:s2_6}
    \end{alignat}
\end{subequations}
The original two-stage OCP \cite{ZSH_24_2s} comprises a first stage with a fixed time grid aligned with the discrete-time system, and a second stage with a variable time grid, of which the total time $T_2$ is minimized for time optimality.
On this basis, the objective (\ref{ocp1:obj}) in the robustified formulation is to minimize the total time $T_2\geq0$ of stage 2 along with the sum of the traces of covariances of states and controls in stage 1, weighted by tunable, positive-definite regularization matrices $R^{\text{regu}}\in\mathbb{R}^{(n_s+n_u)\times (n_s+n_u)}$ and $R^{\text{regu}}_{\text{tf}}\in\mathbb{R}^{n_s\times n_s}$:
\begin{equation}
    \begin{aligned}
    l_{\Sigma}(\Sigma_{1,n}, K_{1,n}) & := \texttt{tr}\left(R^{\text{regu}}\begin{bmatrix}I_{n_s}\\K_{1,n}\end{bmatrix}\Sigma_{1,n}\begin{bmatrix}I_{n_s}\\K_{1,n}\end{bmatrix}^\top\right),\\
    l_{\Sigma_{\text{tf}}}(\Sigma_{1,N_1}) & :=\texttt{tr}\left(R^{\text{regu}}_{\text{tf}}\Sigma_{1,N_1}\right).
    \end{aligned}
    \label{eq: ocp1_obj_cov}
\end{equation}
Constraints (\ref{ocp1:s1_1}-\ref{ocp1:s1_5}) define stage 1 with a horizon length $N_1$, constraints (\ref{ocp1:s2_1}-\ref{ocp1:s2_6}) define stage 2 with a horizon length $N_2$, and the constraint (\ref{ocp1:s1_s2}) stitches the last nominal state $\Bar{s}_{1,N_1}$ of stage 1 to the first nominal state $\Bar{s}_{2,0}$ of stage 2.
In stage 1, we optimize the nominal state and control trajectory \(\Bar{\svector{s}}_1\) and \(\Bar{\svector{u}}_1\), feedback gains \(\smatrix{K}_1\), corresponding state covariances \(\smatrix{\Sigma}_1\), and constraint variances \(\svector{\beta}_1\). 
In stage 2, we focus only on optimizing the nominal state and control trajectory \(\Bar{\svector{s}}_2\) and \(\Bar{\svector{u}}_2\), along with the total time $T_2$ of stage 2.
Note that the constraint (\ref{ocp1:s2_1}) is obtained by discretizing the scaled continuous-time system $\frac{\text{d}\bar{s}(t)}{\text{d}\tau}=f_c(\bar{s}(t), \bar{u}(t))\frac{T_2}{N_2},\ \tau\in[0,N_2]$, where $f_c(\bar{s}(t), \bar{u}(t))$ is the nominal continuous-time equivalence of system (\ref{eq: discrete_time_sys}), and a time scaler $\tau:=N_2t/T_2$ makes the total time $T_2$ independent of the numerical integration.
Constraints (\ref{ocp1:s2_3}) and (\ref{ocp1:s2_5}) of stage 2 are derived from $(K_{1,N_1-1},\Sigma_{1,N_1-1})$ and $\Sigma_{1,N_1}$, respectively, for constraints robustification.
It allows the objective of stage 1 to also contribute to time optimality by, for example,  reducing safety margins on collision avoidance constraints to shorten the moving trajectory, and loosening control limits to allow for greater nominal control.
Lastly, a small offset $\epsilon>0$ is added under each square root in constraints (\ref{ocp1:s1_4}, \ref{ocp1:s2_2}, \ref{ocp1:s2_4}) to ensure differentiability at all feasible points \cite{Messerer_21}.
\begin{algorithm}[b]
    \caption{Timely replanning by integrating ASAP-MPC}
    \label{alg: ocp_asap_mpc}
    \begin{algorithmic}[1]
    \Require $s_{\text{t0}}, \Sigma_{\text{t0}}, N_1, N_2, t_s, n_{\text{update}} \gets N_1, t_{\text{curr}}:=0$
        \State $\Bar{\svector{s}}_1, \Bar{\svector{u}}_1, \smatrix{K}_1, \smatrix{\Sigma}_1, T_2 \gets$ \texttt{tailored\_solver}($s_{\text{t0}}, \Sigma_{\text{t0}}$)\label{alg1_init_plan}
        \State $\Bar{\svector{s}}^\text{exec}, \Bar{\svector{u}}^\text{exec}, \smatrix{K}^\text{exec}\gets\Bar{\svector{s}}_1, \Bar{\svector{u}}_1, \smatrix{K}_1$ \label{alg1_init_exec}
        \Repeat $\:$
        \State $s_{\text{t0}}, \Sigma_{\text{t0}} \gets \Bar{s}_{1,n_{\text{update}}}, \Sigma_{1,n_{\text{update}}}$ \label{alg1_update_init}
        \State start \texttt{tailored\_solver}($s_{\text{t0}}, \Sigma_{\text{t0}}, \text{p}_{\text{OCP}}$) \label{alg1_replan}
        \While{waiting for solution} for $n \in \left[ 0, N_1-1\right]$: \label{alg1_cl_start}
        \State measure $s_{\text{curr}} \gets s(t_{\text{curr}})$
        \State apply $u_{\text{curr}} = \Bar{u}_{n}^\text{exec} + K_{n}^\text{exec}(s_{\text{curr}} - \Bar{s}_{n}^\text{exec})$
        \State $t_{\text{curr}}, n \gets t_{\text{curr}} + t_s, n+1$
        \EndWhile \label{alg1_cl_end}
        \State $\Bar{\svector{s}}_1, \Bar{\svector{u}}_1, \smatrix{K}_1, \smatrix{\Sigma}_1, T_2 \gets$ \texttt{tailored\_solver} solved \label{alg1_replan_done}
        \State get computation time $t_{\mathrm{comp}}$\label{alg1_t_comp}
        \State $n_{\text{update}} \gets \text{ceil}(t_{\mathrm{comp}}/t_s)$\label{alg1_n_update}
        \State Update $\Bar{\svector{s}}^\text{exec}, \Bar{\svector{u}}^\text{exec}, \smatrix{K}^\text{exec}$\label{alg1_update_exec}
        \Until{$T_2 - n_{\mathrm{update}}t_s \leq 0$}
    \end{algorithmic}
\end{algorithm}
\subsection{Timely replanning strategy}\label{Sec:IID}
The two-stage approach provides a low, constant number of control steps, leading to a low and steady computational load \cite{ZSH_24_2s}. 
By integrating ASAP-MPC \cite{ASAP-MPC}, an asynchronous NMPC update strategy, feasibility and timely replanning are ensured by continuously stitching the replanned stage 1 trajectory to the previously planned stage 1 trajectory, assuming the presence of an accurate trajectory tracking controller.

We integrate ASAP-MPC strategy with our robustified two-stage OCP (\ref{ocp: time_optimal_2s_robustified}), eliminating the need for a separate tracking controller assumption, as it is now embedded within the replanning process.
As detailed in Algo. \ref{alg: ocp_asap_mpc}, after obtaining the initial planning result (line \ref{alg1_init_plan}), replanning starts immediately (line \ref{alg1_replan}) with updated initial conditions selected from the current \(\Bar{\svector{s}}_1\) and \(\smatrix{\Sigma}_1\) (line \ref{alg1_update_init}).
Meanwhile, before the replanning result is obtained, the current nominal state and control trajectory \(\Bar{\svector{s}}_1\) and \(\Bar{\svector{u}}_1\), along with the feedback gains \(\smatrix{K}_1\) of stage 1, are the first executables (line \ref{alg1_init_exec}) and are applied in the closed loop (lines \ref{alg1_cl_start}–\ref{alg1_cl_end}).
The replanning result is designed to be obtained within the stage 1 horizon, $N_1t_s$, 
After obtaining the replanning result (line \ref{alg1_replan_done}), the initial conditions are updated with the newly computed \(\Bar{\svector{s}}_1\) and \(\Bar{\svector{u}}_1\) based on the computation time (lines \ref{alg1_t_comp}-\ref{alg1_n_update}), enabling a new replanning to start immediately. 
This characteristic leads us to describe it as \textit{timely replanning}, and a tailored solver is proposed to efficiently solve the OCP (\ref{ocp: time_optimal_2s_robustified}) at lines \ref{alg1_init_plan} and \ref{alg1_replan}, as detailed in Sec. \ref{Sec:III}.
Additionally, the algorithm updates the executables (line \ref{alg1_update_exec}) by seamlessly integrating the remaining portion of the current executables with the first $n_\text{update}$ segment of the newly computed \(\Bar{\svector{s}}_1\), \(\Bar{\svector{u}}_1\), and \(\smatrix{K}_1\). For example, \(\Bar{\svector{s}}^\text{exec} = [\Bar{s}^\text{exec}_{n_\text{update}}, \dots, \Bar{s}^\text{exec}_{N_1-1}, \Bar{s}^1_0, \dots, \Bar{s}^1_{n_\text{update}-1}]\).

The algorithm continues timely replanning until $T_2 - n_{\mathrm{update}}t_s \leq 0$; in other words, until the estimated total time of stage 2 for the next replanning is less than zero, indicating that the system can then reach the terminal state within the stage 1 horizon.
At this point, the final replanning step addresses a simplified version of OCP (\ref{ocp: time_optimal_2s_robustified}), excluding stage 2.
Instead, it incorporates an exponentially weighted objective to achieve time optimality \cite{Verschueren_17}, as follows:
\begin{subequations} 
    \label{ocp: robust_exp}
    \begin{alignat}{3}
        \minimize_{\substack{
        \Bar{\svector{s}}, \Bar{\svector{u}},\\
        \smatrix{K},\smatrix{\Sigma}, 
        \svector{\beta}
        }} 
        \sum_{n=0}^{N-1}\gamma^n\|\Bar{s}_n-s_{\text{tf}}\|_1&+l_{\Sigma}(\Sigma_{n}, K_{n})+ l_{\Sigma_{\text{tf}}}(\Sigma_{N})\\
        \text{s.t.}\hspace{5em}
        \Bar{s}_{0}   =&\  s_{\text{t0}}, \Sigma_{0} = \Sigma_{\text{t0}},\\
        \Bar{s}_{n+1} =&\ f_d(\Bar{s}_{n},\Bar{u}_{n}),\\
        \Sigma_{n+1}  =&\ \Phi(\Bar{s}_{n},\Bar{u}_{n}, K_{n}, \Sigma_{n}),\\
        h(\Bar{s}_{n}, \Bar{u}_{n}) &+ \sigma\sqrt{\beta_{n}+\epsilon}\leq0, \\
        \beta_{n}     =&\ H(\Bar{s}_{n},\Bar{u}_{n}, K_{n}, \Sigma_{n}),\\
        h_{\text{tf}}(\Bar{s}_{N}) &+\sigma\sqrt{\beta_{N} +\epsilon}\leq0,\\
        \beta_{N}   =&\ H_{\text{tf}}(\Bar{s}_{N}, \Sigma_{N}),\\
        \Bar{s}_{N}   =&\  s_{\text{tf}}, 
    \end{alignat}
\end{subequations}
where $\gamma$ is an exponential weighting factor greater than 1, but it should not be set too high to avoid numerical ill-conditioning \cite{ZSH_24_2s}.

%% file: SecIII.tex
\section{Tailored algorithm to solve the robustified two-stage time-optimal OCP}\label{Sec:III}
In this section, we detail \texttt{tailored\_solver} presented in Algo.\ref{alg: ocp_asap_mpc} to efficiently solve OCP (\ref{ocp: time_optimal_2s_robustified}) given initial conditions $s_{\text{t0}}, \Sigma_{\text{t0}}$, which is too intricate to be tackled directly by a generic solver like Ipopt \cite{ipopt}. 
Inspired by \cite{Messerer_21}, we decouple OCP (\ref{ocp: time_optimal_2s_robustified}) into two subproblems: a Riccati recursion to derive feedback gains based on the nominal trajectory and corresponding dual variables, and a nominal time-optimal OCP with safety margins derived from these feedback gains. 
The algorithm iterates by alternately solving these two subproblems, which helps reduce the solving time of OCP (\ref{ocp: time_optimal_2s_robustified}), even with the large number of decision variables requiring optimization.

We rewrite OCP (\ref{ocp: time_optimal_2s_robustified}) as the following compact form:
\begin{subequations} \label{ocp: compact}
    \begin{alignat}{3}
        \minimize_{\substack{
        T_2, \svector{z}, \smatrix{M}, \svector{\beta}
        }} 
        &T_2 + L(\svector{z}, \smatrix{M})\label{ocp2:obj}\\
        \text{s.t.}\hspace{1em} &g(\svector{z}, T_2)=0, \label{ocp2:1}\\
        &h(\svector{z})+\sigma\sqrt{\svector{\beta}+\epsilon}\leq0,\label{ocp2:2}\\
        &H(\svector{z}, \smatrix{M}) - \svector{\beta} = 0,\label{ocp2:3}\\
        &-T_2\leq0,
    \end{alignat}
\end{subequations}
where $\smatrix{M}\in\mathbb{R}^{n_M}:=\texttt{vec}(\texttt{vec}(K_{1,0}), ..., \texttt{vec}(K_{1,N_1-1}))$ comprises vectorized $\smatrix{K}_1$,  $\svector{z}\in\mathbb{R}^{n_z}:=\texttt{vec}(\Bar{\svector{s}}_1, \Bar{\svector{u}}_1, \Bar{\svector{s}}_2, \Bar{\svector{u}}_2)$ contains nominal state and control trajectory of stage 1 and 2. 
The state covariances $\smatrix{\Sigma}_1$ have been eliminated from the OCP (\ref{ocp: time_optimal_2s_robustified}), while $\svector{\beta}\in\mathbb{R}^{n_\beta}$, containing both $\svector{\beta}_1$ and $\svector{\beta}_2$, is preserved for utilization in the proposed tailored algorithm.
The equality constraints (\ref{ocp2:1}) and (\ref{ocp2:3}) summarizes (\ref{ocp1:s1_1},\ref{ocp1:s1_2},\ref{ocp1:s1_s2},\ref{ocp1:s2_1},\ref{ocp1:s2_6}), and (\ref{ocp1:s1_5}, \ref{ocp1:s2_3}, \ref{ocp1:s2_5}), respectively.
The inequality constraint (\ref{ocp2:2}) summarizes (\ref{ocp1:s1_4}, \ref{ocp1:s2_2}, \ref{ocp1:s2_4}).

The Lagrangian of (\ref{ocp: compact}) is given by
\begin{equation*}
\begin{aligned}
    &\mathcal{L}(T_2, \svector{z}, \smatrix{M}, \svector{\beta}, \svector{\lambda}, \svector{\mu}, \svector{\eta}, \rho) = T_2 + L(\svector{z}, \smatrix{M}) + \svector{\lambda}^\top g(\svector{z}, T_2) \\
    &+ \svector{\mu}^\top(h(\svector{z})+\sigma\sqrt{\svector{\beta}+\epsilon}) + \svector{\eta}^\top(H(\svector{z}, \smatrix{M})-\svector{\beta})-\rho T_2,
\end{aligned}
\label{eq: Lagrangian}
\end{equation*}
with dual variables $\svector{\lambda}, \svector{\mu}, \svector{\eta}, \rho$. 
The KKT conditions for (\ref{ocp: compact}) are as follows:
\begin{equation}\label{eq: KKT}
    \begin{aligned}
        \nabla_{T_2}(T_2+\svector{\lambda}^\top g(\svector{z},T_2)-\rho T_2)) &=0,\\
        \nabla_{\svector{z}}(L(\svector{z}, \smatrix{M})+\svector{\lambda}^\top g(\svector{z},T_2)+\svector{\mu}^\top h(\svector{z})+\svector{\eta}^\top H(\svector{z}, \smatrix{M})) &=0,\\
        \nabla_{\smatrix{M}}(L(\svector{z}, \smatrix{M}) + \svector{\eta}^\top H(\svector{z}, \smatrix{M})) &=0,\\
        \nabla_{\svector{\beta}}\mathcal{L}:\ \forall \beta_i,\mu_i,\eta_i\in\mathbb{R}, i\in[1,n_\beta], \frac{\mu_i\sigma}{2\sqrt{\beta_i+\epsilon}}-\eta_i &=0,\\
        \nabla_{\svector{\lambda}}\mathcal{L}:\ g(\svector{z}, T_2)&=0,\\
        0\leq\svector{\mu} \perp \nabla_{\svector{\mu}}\mathcal{L}:\ h(\svector{z})+\sigma\sqrt{\svector{\beta}+\epsilon}&\leq0,\\
        \nabla_{\svector{\eta}}\mathcal{L}:\ H(\svector{z}, \smatrix{M})-\svector{\beta}&=0,\\
        0\leq\rho\perp \nabla_{\rho}\mathcal{L}:\ -T_2&\leq0,
    \end{aligned}
\end{equation}
where $n_\beta:=(N_1+N_2)n_h+n_{h_{\text{tf}}}$.

We decompose OCP (\ref{ocp: compact}), along with its corresponding KKT conditions (\ref{eq: KKT}), into two subproblems. 
The first subproblem optimizes feedback gains $\smatrix{K}_1$ given fixed $\Bar{\svector{z}}, \Bar{T}_2, \Bar{\svector{\eta}}$ with the KKT condition  
\begin{equation}\label{eq: KKT_K}
    \begin{aligned}
        \nabla_{\smatrix{M}}(L(\Bar{\svector{z}}, \smatrix{M}) + \Bar{\svector{\eta}}^\top H(\Bar{\svector{z}}, \smatrix{M})) &=0.\\
    \end{aligned}
\end{equation}
The second subproblem optimizes $T_2$ and the nominal trajectory $\svector{z}$ by defining a nominal time-optimal problem with uncertainty terms, i.e., terms related to feedback gains, fixed.
Following is the corresponding KKT conditions
\begin{equation}\label{eq: KKT_nominal}
    \begin{aligned}
        \nabla_{T_2}(T_2+\svector{\lambda}^\top g(\svector{z},T_2)-\rho T_2) &=0,\\
        \nabla_{\svector{z}}(L(\Bar{\svector{z}}, \Bar{\smatrix{M}})+\svector{\lambda}^\top g(\svector{z},T_2)+\svector{\mu}^\top h(\svector{z})+\Bar{\svector{\eta}}^\top H(\Bar{\svector{z}}, \Bar{\smatrix{M}})) &=0,\\
        \nabla_{\svector{\lambda}}\mathcal{L}:\ g(\svector{z}, T_2)&=0,\\
        0\leq\svector{\mu} \perp \nabla_{\svector{\mu}}\mathcal{L}:\ h(\svector{z})+\sigma\sqrt{H(\Bar{\svector{z}}, \Bar{\smatrix{M}})+\epsilon}&\leq0,\\
        0\leq\rho\perp \nabla_{\rho}\mathcal{L}:\ -T_2&\leq0.
    \end{aligned}
\end{equation}
Note that $\bar{\svector{\beta}}=H(\bar{\svector{z}}, \bar{\smatrix{M}})$ can be determined for fixed $\bar{\svector{z}}, \bar{\smatrix{M}}$.
The corresponding dual variable $\bar{\svector{\eta}}$ is obtained, given $\bar{\svector{\mu}}$ after solving the second subproblem, as follow
\begin{equation}
    \Bar{\eta}_i = \frac{\Bar{\mu}_i\sigma}{2\sqrt{\Bar{\beta}_i+\epsilon}}, i\in[1,n_\beta].
    \label{eq: mu_to_eta}
\end{equation}
\subsection{Subproblem: Optimize Feedback Gains}\label{Sec: IIIA}
The subproblem to optimize feedback gains with the KKT condition (\ref{eq: KKT_K}) is
\begin{subequations} \label{ocp: optimize_K}
    \begin{alignat}{3}
        \minimize_{\substack{
        \smatrix{\Sigma}_1, \smatrix{K}_1
        }} 
        \sum_{n=0}^{N_1-1}&\Tilde{l}_{\Sigma}(\Sigma_{1,n}, K_{1,n})
        + \Tilde{l}_{\Sigma_{\text{tf}}}(\Sigma_{1,N_1})\\
        \text{s.t.}\hspace{1em} \Sigma_{1,0}&= \Sigma_{\text{t0}},\\
        \Sigma_{1,n+1}&=\Phi\left(\Bar{s}_{1,n}, \Bar{u}_{1,n}, K_{1,n}, \Sigma_{1,n}\right),
    \end{alignat}
\end{subequations}
where $\Tilde{l}_{\Sigma}$ and $\Tilde{l}_{\Sigma_{\text{tf}}}$ are same as (\ref{eq: ocp1_obj_cov}) but with regularization terms $R_{1,n}:=R^{\text{regu}} + R^h_{1,n}$ and $R_{\text{tf}}:=R^{\text{regu}}_{\text{tf}} + R^h_{\text{tf}}$.
Therein,
\begin{equation*}
    R^h_{\text{tf}} =\frac{\partial h_{\text{tf}}(\bar{s}_{2,N_2})}{\partial s_{2,N_2}}^\top\texttt{diag}(\bar{\eta}_{2,N_2})\frac{\partial h_{\text{tf}}(\bar{s}_{2,N_2})}{\partial s_{2,N_2}},
\end{equation*}
for $0\leq n < N_1-1$, 
\begin{equation*}
    R^h_{1,n} = \frac{\partial h(\bar{s}_{1,n}, \bar{u}_{1,n})}{\partial \texttt{vec}(s_{1,n}, u_{1,n})}^\top\texttt{diag}(\bar{\eta}_{1,n})\frac{\partial h(\bar{s}_{1,n}, \bar{u}_{1,n})}{\partial \texttt{vec}(s_{1,n}, u_{1,n})},
\end{equation*}
and for $n = N_1-1$,
\begin{equation*}
\begin{aligned}
    &R^h_{1,n} = \frac{\partial h(\bar{s}_{1,n}, \bar{u}_{1,n})}{\partial \texttt{vec}(s_{1,n}, u_{1,n})}^\top\texttt{diag}(\bar{\eta}_{1,n})\frac{\partial h(\bar{s}_{1,n}, \bar{u}_{1,n})}{\partial \texttt{vec}(s_{1,n}, u_{1,n})}\\
    +&\sum_{m=0}^{N_2-1}\frac{\partial h(\bar{s}_{2,m}, \bar{u}_{2,m})}{\partial \texttt{vec}(s_{2,m}, u_{2,m})}^\top\texttt{diag}(\bar{\eta}_{2,m})\frac{\partial h(\bar{s}_{2,m}, \bar{u}_{2,m})}{\partial \texttt{vec}(s_{2,m}, u_{2,m})},
\end{aligned}
\end{equation*}
which also takes into account activated constraints of stage 2.

Note that $\Sigma_{\text{t0}}$ and $\Bar{\svector{z}}, \Bar{T}_2, \Bar{\svector{\eta}}$ are given, i.e., from solving the nominal time-optimal problem presented in Sec. \ref{Sec: IIIB}.
Split $R_{1,n}$ into block components $R_{1,n}=\begin{bmatrix}R_{1,n}^s &R_{1,n}^{su}\\{R_{1,n}^{su}}^\top&R_{1,n}^u\end{bmatrix}$.
Since $R_{1,n}\succeq0$, $R_{\text{tf}}\succeq0$, $R_{1,n}^s\succeq0$, $R_{1,n}^u\succ0$,
following \cite[Lemma 11]{Messerer_21}, the subproblem (\ref{ocp: optimize_K}) can be rewritten as an unconstrained problem, whose solution is uniquely defined by the Riccati recursion
\begin{equation}
    \begin{aligned}
        S_{N_1}&=R_{\text{tf}},\\
        K_{n}^\ast&=-(R_{n}^{u}+B_{n}^\top S_{n+1}B_{n})^{-1}({R_{n}^{su}}^\top+B_{n}^\top S_{n+1}A_{n}),\\
        S_{n}&=R_{n}^{s}+A_n^\top S_{n+1}A_n+(R_{n}^{su}+A_n^\top S_{n+1}B_{n})K_{n}^\ast,
    \end{aligned}
    \label{eq: riccati}
\end{equation}
where $A_n$ and $B_{n}$ refers to (\ref{eq: uncertainty_propagation}) with the stage index omitted for brevity.
\subsection{Subproblem: Optimize Nominal Time-optimal OCP}\label{Sec: IIIB}
The second subproblem, governed by the KKT conditions (\ref{eq: KKT_nominal}), involves deriving the nominal state and control trajectory as well as the minimal total time of stage 2 by solving the following nominal two-stage time-optimal OCP with constraint uncertainties frozen:
\begin{subequations} \label{ocp: time_optimal_nominal}
    \begin{alignat}{3}
        \minimize_{
        \Bar{\svector{s}}_1, \Bar{\svector{u}}_1, \Bar{\svector{s}}_2, \Bar{\svector{u}}_2, T_2
        } 
        \hspace{1em}&\substack{T_2+\Bar{c}^\top\text{vec}(\Bar{\svector{s}}_1, \Bar{\svector{u}}_1, \Bar{\svector{s}}_2, \Bar{\svector{u}}_2)}\label{ocp3:obj}\\
        \text{s.t.}\hspace{1em}
        \Bar{s}_{1,0}   =&\  s_{\text{t0}}\label{ocp3:s1_1}\\
        \Bar{s}_{1,n+1} =&\ f_d(\Bar{s}_{1,n},\Bar{u}_{1,n}),\label{ocp3:s1_2}\\
        h(\Bar{s}_{1,n}, \Bar{u}_{1,n}) &+ \Bar{h}^{\text{sm}}_{1,n}\leq0, \label{ocp3:s1_3}\\
        \Bar{s}_{2,0}   =&\  \Bar{s}_{1,N_1},\label{ocp3:s1_s2}\\
        \Bar{s}_{2,n+1} =&\ f_T(\Bar{s}_{2,n},\Bar{s}_{2,n},T_2/N_2),\label{ocp3:s2_1}\\
        h(\Bar{s}_{2,n},\Bar{u}_{2,n}) &+ \Bar{h}^{\text{sm}}_{2,n}\leq0,\label{ocp3:s2_2}\\ 
        h_{\text{tf}}(\Bar{s}_{2,N_2}) &+ \Bar{h}^{\text{sm}}_{\text{tf}}\leq0.\label{ocp3:s2_3}
    \end{alignat}
\end{subequations}
After obtaining feedback gains $\smatrix{K}_1$ by solving the subproblem (\ref{ocp: optimize_K}) given $\Bar{\svector{z}}, \Bar{T}_2, \Bar{\svector{\eta}}$ from previous iteration, the state covariance $\smatrix{\Sigma}_1$ of stage 1 is propagated via (\ref{eq: uncertainty_propagation}).
Safety margin terms $\Bar{h}^{\text{sm}}_{1,n}, \Bar{h}^{\text{sm}}_{2,n}, \Bar{h}^{\text{sm}}_{\text{tf}}:=\sigma\sqrt{\bar{\beta}+\epsilon}$ are then updated with $\bar{\beta}$ obtained via (\ref{eq: constrs_var}) for resolving the OCP (\ref{ocp: time_optimal_nominal}).
Additionally, the objective (\ref{ocp3:obj}) includes gradient correction terms $\Bar{c}^\top\text{vec}(\Bar{\svector{s}}_1, \Bar{\svector{u}}_1, \Bar{\svector{s}}_2, \Bar{\svector{u}}_2)$ to satisfy the KKT conditions (\ref{eq: KKT_nominal}).
The correction coefficients $\Bar{c}$ are derived from $\nabla_{\svector{z}}L(\Bar{\svector{z}}, \Bar{\smatrix{M}}) + \nabla_{\svector{z}} H(\Bar{\svector{z}}, \Bar{\smatrix{M}}, \Bar{T}_2)\Bar{\svector{\eta}}$. 

Algo.\ref{alg: ocp_solver} summarizes the process of alternately solving two subproblems and the sequence for updating all optimization variables and coefficients. 
In terms of implementation, to facilitate computation, steps to update $\bar{\svector{\eta}}, \smatrix{K}_1, \smatrix{\Sigma}_1, \bar{\svector{\beta}}$, safety margins and gradient corrections can be symbolically formulated in advance.
Solving (\ref{ocp: time_optimal_nominal}) in step \ref{alg2_resolve_ocp} can be warm-started using the solutions from the previous iteration.
\begin{algorithm}[t]
    \caption{Tailored solver}
    \label{alg: ocp_solver}
    \begin{algorithmic}[1]
    \Require $s_{\text{t0}}, \Sigma_{\text{t0}}, \Bar{h}^{\text{sm}}_{\text{init}}, \Bar{c}_{\text{init}}=0$, 
    \State $\Bar{\svector{s}}_1, \Bar{\svector{u}}_1, \Bar{\svector{s}}_2, \Bar{\svector{u}}_2, T_2\gets$ solve the initial iteration of (\ref{ocp: time_optimal_nominal})
    \While{true} $\:$
    \State $\Bar{\svector{\eta}}\gets$ obtained via (\ref{eq: mu_to_eta})
    \State $\smatrix{K}_1\gets$ obtained via (\ref{eq: riccati})
    \State $\smatrix{\Sigma}_1\gets$ obtained via (\ref{eq: uncertainty_propagation})
    \If{KKT conditions (\ref{eq: KKT}) meet threshold}
    \State output $\Bar{\svector{s}}_1, \Bar{\svector{u}}_1, \smatrix{K}_1, \smatrix{\Sigma}_1, T_2$
    \State break
    \EndIf
    \State $\Bar{\svector{\beta}}\gets$ obtained via (\ref{eq: constrs_var})
    \State $\Bar{h}^{\text{sm}}_{1,n}, \Bar{h}^{\text{sm}}_{2,n}, \Bar{h}^{\text{sm}}_{\text{tf}}\gets$ updated via (\ref{eq: constrs_var})
    \State $\Bar{c}\gets$ updated via $\nabla_{\svector{z}}L(\Bar{\svector{z}}, \Bar{\smatrix{M}}) + \nabla_{\svector{z}} H(\Bar{\svector{z}}, \Bar{\smatrix{M}}, \Bar{T}_2)\Bar{\svector{\eta}}$.
    \State $\Bar{\svector{s}}_1, \Bar{\svector{u}}_1, \Bar{\svector{s}}_2, \Bar{\svector{u}}_2, T_2\gets$ solve (\ref{ocp: time_optimal_nominal}) with updated safety margins and gradient corrections \label{alg2_resolve_ocp}
    \EndWhile
    \end{algorithmic}
\end{algorithm}

%% file: SecIV.tex
\section{Numerical Example and Discussions}\label{Sec:IV}
We now demonstrate and discuss the performance of the proposed robustified two-stage time-optimal OCP, along with the corresponding timely replanning and solving algorithms, using a numerical example. 
This evaluation examines aspects such as time optimality, the handling of uncertainties and safety margins, and the computation time by comparing the results of timely replanning with those of solving a single planning problem.

The OCPs are formulated in Python using the Rockit Toolbox \cite{rockit}, which is developed for rapid OCP prototyping.
Ipopt \cite{ipopt} with the ma57 linear solver \cite{hsl} is employed to solve the OCP. 
Additional algorithmic steps, including Riccati recursion and uncertainty propagation, are implemented using CasADi \cite{casadi}.
All computations are conducted on a laptop equipped with an Intel$^{\circledR}$ Core\textsuperscript{\texttrademark} i7-1185G7 processor with eight cores at 3GHz and 31.1GB RAM.

\begin{table}[b]
\caption{\small{parameter values}}
\centering
\tablefontsize
\begin{tabular}{P{4em} P{4em} P{4em} P{4em} P{3em} P{3em}}
    \toprule
    $t_{s}$ & 
    $v_{\text{min}/\text{max}}$ & 
    $\omega_{\text{min}/\text{max}}$ &
    $N_1(N_2)$&
    $\sigma$&
    $\epsilon$
    \\  
    \midrule
    0.02s & 
    0/0.5m/s &
    $\pm\frac{\pi}{4}$$\text{rad}/\text{s}$ &
    30&
    3&
    1e-8
    \\
    \bottomrule
\end{tabular}
\label{table}
\end{table}
This example involves transitioning a mobile robot, modelled as unicycle $\Dot{s}=[v\cos\theta, v\sin\theta, \omega]^\top$ with $n_s=3$ states ($x,y$ positions and heading angle $\theta$) and $n_u=2$ control inputs (forward velocity $v$ and angular velocity $\omega$), from its initial state $s_{\text{t0}}=[0.1\text{m}, 0.5\text{m}, 0\text{rad}]^\top$ without uncertainty, i.e., $\Sigma_{\text{t0}}=0$, to a terminal state $s_{\text{tf}}=[2.5\text{m}, 1\text{m}, 0\text{rad}]^\top$ in the minimal time, while also avoiding collision with an ellipsoidal obstacle, defined as $h_{\text{e}}: 1-{p^{\text{diff}}}^\top\Omega_{\text{e}}p^{\text{diff}}\leq0$, where $p^{\text{diff}}:=\begin{bmatrix}x-x_e\\y-y_e\end{bmatrix}$, and $\Omega_{\text{e}}:=R(\theta_e)^\top\text{diag}(\frac{1}{a_e^2}, \frac{1}{b_e^2})R(\theta_e)$ with $R(\theta_e)$ representing the elliptical rotation matrix, and parameters $[x_e,y_e,a_e,b_e,\theta_e] = [1.25\text{m}, 0.5\text{m}, 1\text{m}, 0.5\text{m}, \pi/6\text{rad}]$.
An additive process noise in discrete time is considered with covariance $\Sigma_w=10^{-6}\text{diag}((1\text{m})^2, (1\text{m})^2, (1.75\text{rad})^2)$.

To solve this example using timely replanning by executing Algo.\ref{alg: ocp_asap_mpc}, regularization matrices in OCP (\ref{ocp: time_optimal_2s_robustified}) are set as $R^{\text{regu}}=I_5$ and $R^{\text{regu}}_{\text{tf}}=50I_3$, applying larger penalization on the final state covariance to encourage the system to reach the terminal state with low uncertainty.
For the final replanning step, executed after Algo.\ref{alg: ocp_asap_mpc} meets the condition  $T_2 - n_{\mathrm{update}} t_s \leq 0$  and solves OCP (\ref{ocp: robust_exp}), the same regularization matrices are used, with a horizon length of  $N = N_1 + N_2$  and an exponential weighting factor of  $\gamma = 1.015$.
Refer to Table \ref{table} for values of remaining parameters.
Note that the stage 1 horizon, $N_1t_s = 0.6$s, is specifically designed in this example to enable timely replanning during the execution of Algo.\ref{alg: ocp_asap_mpc}.

We compare the results of timely replanning with those of a single planning by solving OCP (\ref{ocp: robust_exp}) with a horizon $N=300$ that is sufficiently long to reach the target from the initial state.
Regularization matrices $R^{\text{regu}}=\texttt{diag}(80,80,80,500,500)$ and $R^{\text{regu}}_{\text{tf}}=1000I_3$ are appropriately chosen to ensure feasibility, with an exponential weighting factor $\gamma=1.015$. 

\begin{figure}[t]
    \centering
    \includegraphics[width=\linewidth]{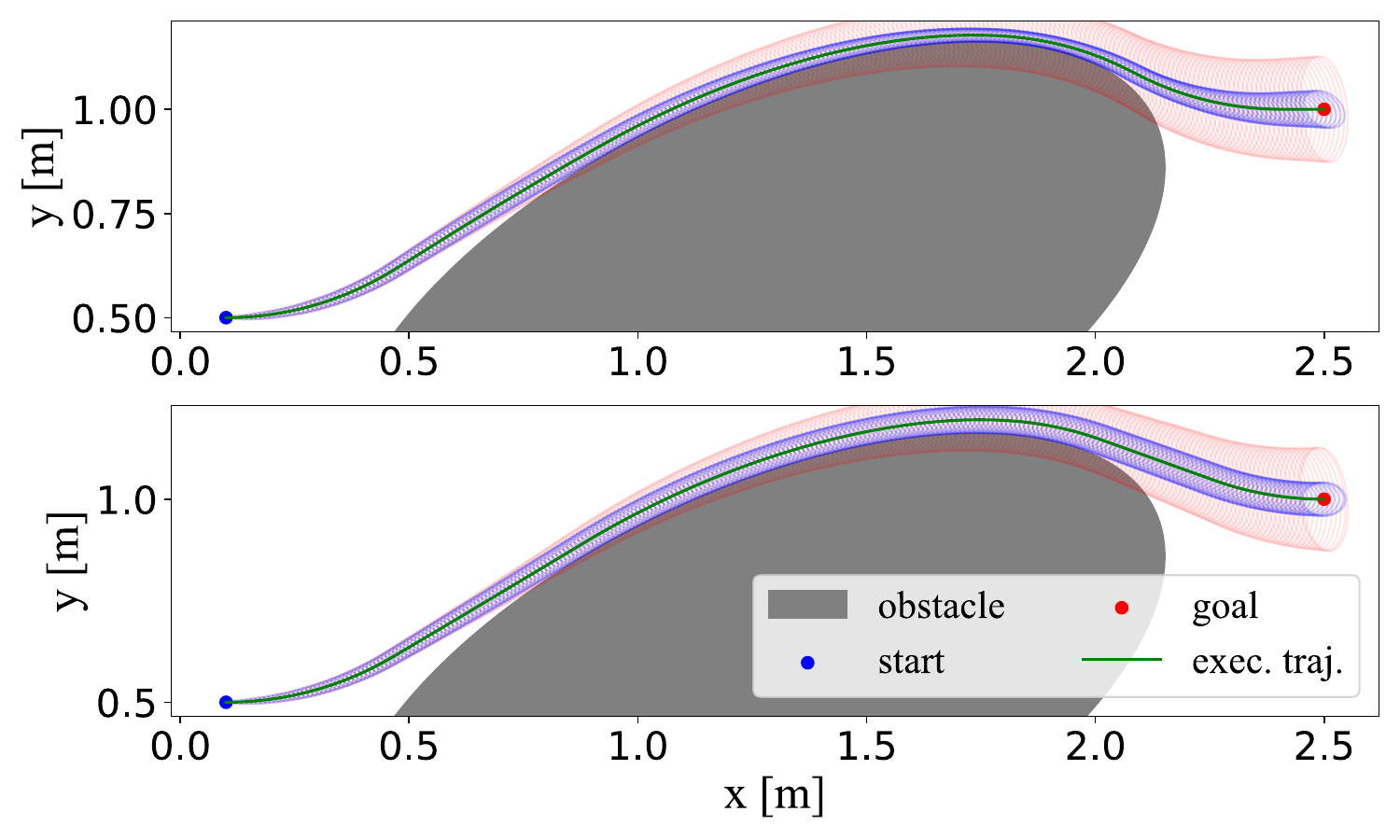}
    \caption{Point-to-point nominal trajectory and the corresponding closed-loop (\blue{blue ellipses})/open-loop (\red{red ellipses}) uncertainty. Top: single planning. Bottom: timely replanning.}
    \label{fig:traj}
\end{figure}

Fig.\ref{fig:traj} shows a comparison of the nominal $x-y$ trajectory and the corresponding closed-loop/open-loop uncertainty on the $x-y$ plane, between single planning and timely replanning.
Both approaches effectively counteract the growth of uncertainty through closed-loop feedback.
The optimal motion times for single planning and timely replanning are 5.2 seconds and 5.22 seconds, respectively, the difference being only one sampling time $t_s$.
We argue that the optimal motion times of the two approaches are not necessarily the same, as they are derived from executing different feedback gains based on distinct problem setting.

\begin{figure}[t]
    \centering
    \includegraphics[width=\linewidth]{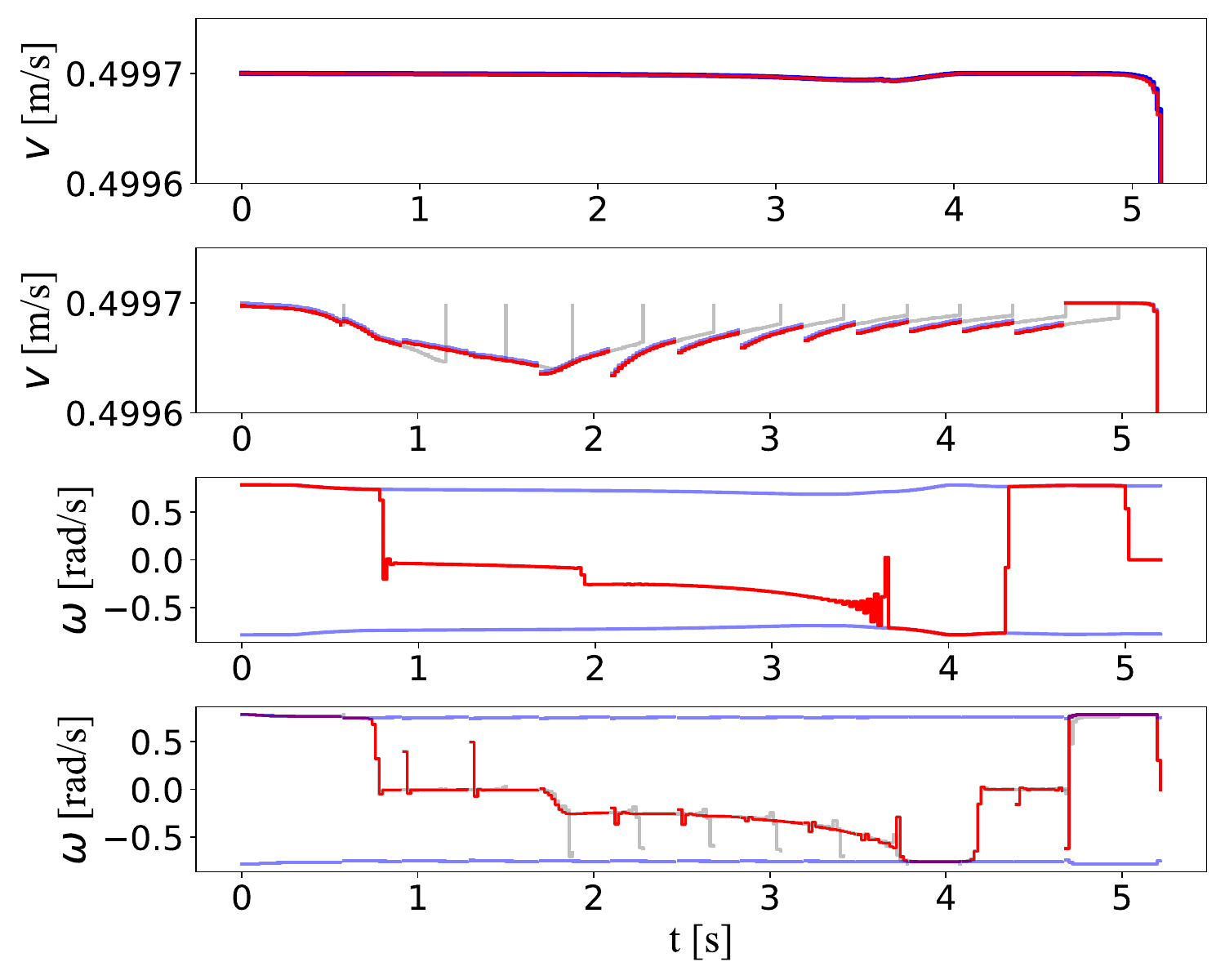}
    \caption{Nominal controls: from top to bottom, linear velocity from single planning and timely replanning, and angular velocity from single planning and timely replanning, respectively. \red{Red} lines denote the executed nominal control trajectory, \blue{blue} lines indicate control limits for the nominal controls considering safety margins, and gray lines represent the remaining planned nominal control trajectory of each replanning.}
    \label{fig:control}
\end{figure}

Fig.\ref{fig:control} shows the control limits, accounting for safety margins, and the resulting nominal controls of both approaches.
Both approaches result in execution at the maximum allowable nominal linear velocities to achieve time optimality.
However, due to different problem setting, both approaches derive distinct safety margins for linear and angular velocities, resulted from different feedback actions on both controls.
Specifically, the single planning approach establishes larger safety margins to allow more feedback actions on angular velocity, particularly between 3 and 4 seconds.
This is illustrated in the second-to-last plot in Fig.\ref{fig:control}, where the blue line rises slightly, creating a larger margin relative to the minimum angular velocity $\omega_\text{min}= -\pi/4$.
In contrast, the timely replanning approach allocates larger safety margins to enhance feedback actions on linear velocity.
Nevertheless, for each replanning between 2 and 4 seconds, it attempts to reduce these safety margins, as shown in the second plot in Fig.\ref{fig:control}, where blue lines are rising.
During this period, as the system moves along the edge of the obstacle, both the regularization terms in the objective (\ref{ocp1:obj}) and the activated collision avoidance constraints in both stages contribute to generating feedback that reduces state and control uncertainty.
Since safety margins on stage 2 constraints are derived from $(K_{1,N_1-1},\Sigma_{1,N_1-1})$ of stage 1, this permits a higher maximum allowable nominal linear velocity in stage 2, also demonstrates that the regularization terms in the objective (\ref{ocp1:obj}) plays a role in achieving overall time optimality.

\begin{figure}[t]
    \centering
    \includegraphics[width=\linewidth]{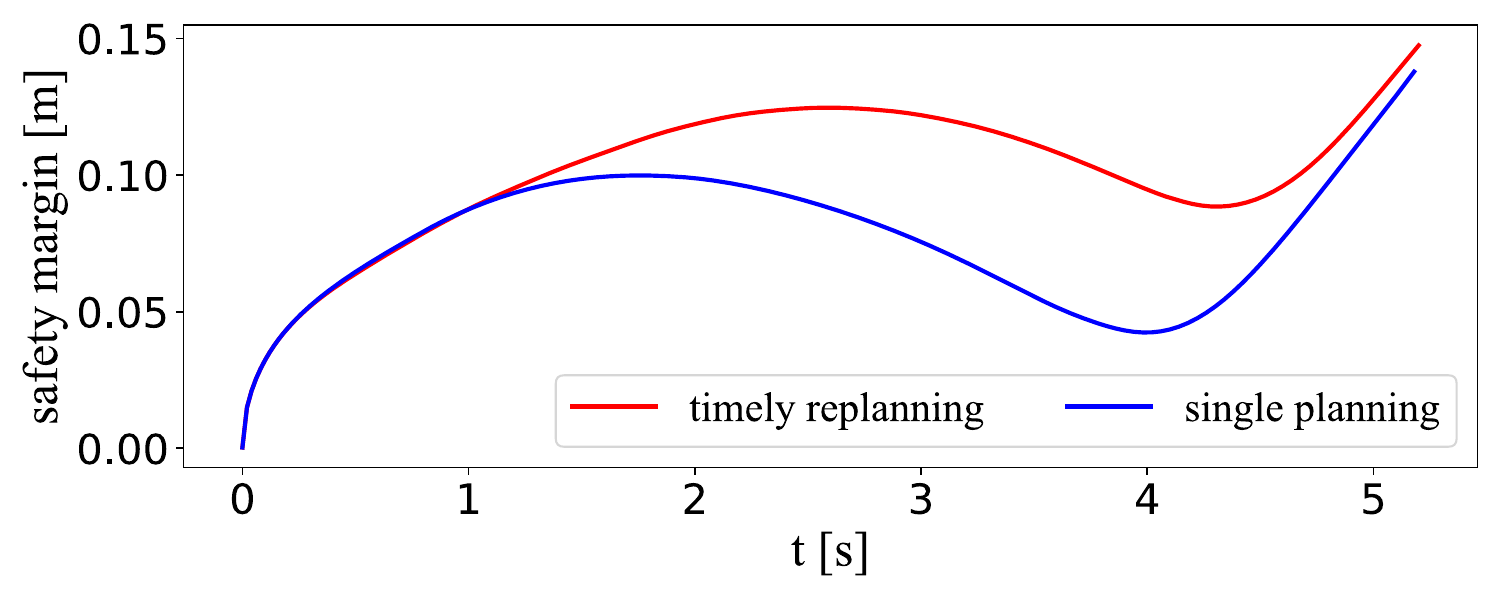}
    \caption{Safety margins on the collision avoidance constraint.}
    \label{fig:ca_sm}
\end{figure}

State uncertainty propagates differently with the feedback actions of each approach, which in turn affects the safety margins on the collision avoidance constraint.
In Fig.\ref{fig:ca_sm}, we compare the collision avoidance safety margins from both approaches. 
It is evident that they follow a similar pattern: as the system moves along the edge of the obstacle between 2 and 4 seconds, the safety margins decrease, indicating that the activated collision avoidance constraints also influence the feedback derived to achieve time optimality.
The difference in collision avoidance safety margins between the two approaches has negligible impact on the nominal trajectory length, which is 2.597 m for single planning and 2.604 m for timely replanning.

\begin{figure}[t]
    \centering
    \includegraphics[width=\linewidth]{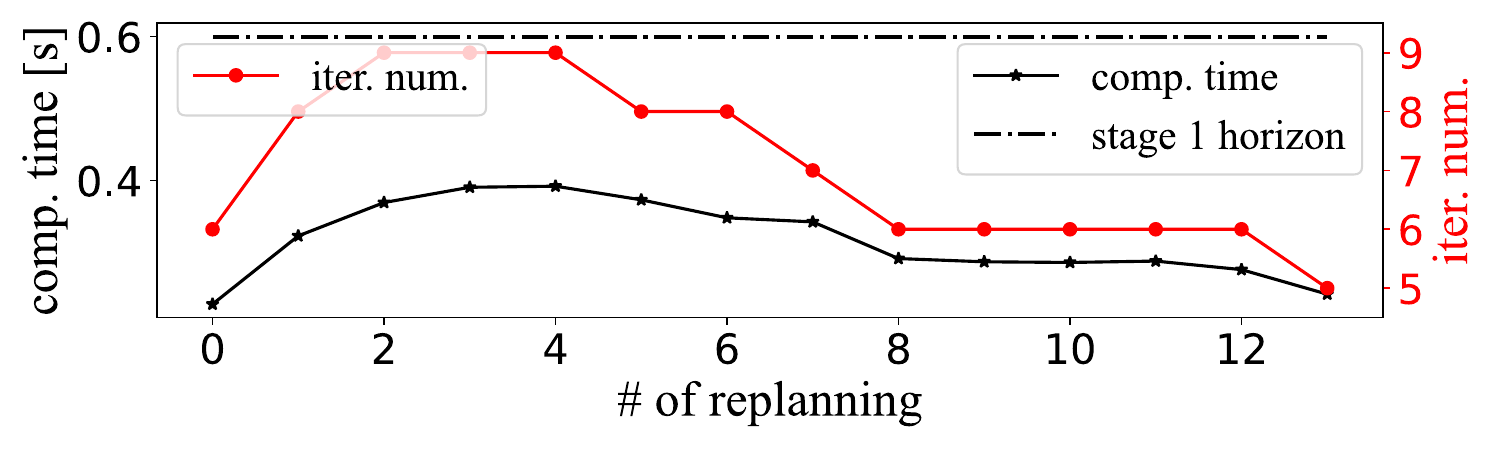}
    \caption{Computation time and iteration number of each replanning.}
    \label{fig:comp_time}
\end{figure}

Lastly, we discuss the computation time. 
For single planning, we set the KKT condition tolerance to $5\times10^{-3}$ and use Algo.\ref{alg: ocp_solver} to solve it. 
It converges after 10 iterations, with a total computation time of 2.1027 seconds.
For timely replanning, we set the KKT condition tolerance to $5\times10^{-5}$.
The computation time and iteration count for each replanning are shown in Fig.\ref{fig:comp_time}.
Each replanning computation is completed within the 0.6-second horizon of stage 1, enabling effective timely replanning and continuous execution of feedback controls.
The results indicate that the timely replanning approach with solving OCP 
(\ref{ocp: time_optimal_2s_robustified}) offers a significant advantage on computation time while achieving similar time-optimal motion performance under uncertainty. 

%% file: SecV.tex
\section{Conclusion}\label{Sec:V}
This paper presents a robustified two-stage OCP to formulate the time-optimal motion planning and control problem under uncertainty, which is modeled as process noise.
The proposed OCP is efficiently solved using a tailored iterative algorithm and is used in an asynchronous NMPC scheme to achieve online replanning.
Future work will focus on optimizing algorithm coding for embedded mobile systems and exploring its real-time potential in an actual AMR for time-optimal tasks.